\documentclass[11pt,a4paper]{article}
\usepackage[hyperref]{emnlp-ijcnlp-2019}
\usepackage[utf8]{inputenc}
\usepackage[T2A,T1]{fontenc}
\usepackage[russian,english]{babel}
\usepackage{times}
\usepackage{latexsym}
\usepackage{multirow}
\usepackage{amsmath}
\usepackage{array, booktabs, adjustbox, multirow, amssymb}
\usepackage{url}
\usepackage[most]{tcolorbox}

\aclfinalcopy 

\title{Mapping (Dis-)Information Flow about the MH17 Plane Crash}

\author{Mareike Hartmann \\
  Dep. of Computer Science\\
  University of Copenhagen \\
  Denmark \\
  {\tt hartmann@di.ku.dk} \\\And
  Yevgeniy Golovchenko \\
   Dep. of Political Science \\
  University of Copenhagen \\
  Denmark \\
  {\tt yg@ifs.ku.dk} \\\And
  Isabelle Augenstein\\
  Dep. of Computer Science\\
  University of Copenhagen \\
  Denmark \\
  {\tt augenstein@di.ku.dk} \\}

\date{}

\begin{document}
\maketitle
\begin{abstract}
Digital media enables not only fast sharing
of information, but also disinformation. One prominent case of an event leading to circulation of disinformation on social media is the MH17 plane crash. Studies analysing the spread of information about this event on Twitter have focused on small, manually annotated datasets, or used proxys for data annotation. In this work, we examine to what extent text classifiers can be used to label data for subsequent content analysis, in particular we focus on predicting pro-Russian and pro-Ukrainian Twitter content related to the MH17 plane crash. Even though we find that a neural classifier improves over a hashtag based baseline, labeling pro-Russian and pro-Ukrainian content with high precision remains a challenging problem. We provide an error analysis underlining the difficulty of the task and identify factors that might help improve classification in future work. Finally, we show how the classifier can facilitate the annotation task for human annotators.
\end{abstract}

\section{Introduction}
Digital media enables fast sharing of information,
including various forms of false or
deceptive information. Hence, besides bringing the obvious advantage of broadening information access for everyone, digital media can also be misused for campaigns that spread disinformation about specific events, or campaigns that are targeted at specific individuals or governments. Disinformation, in this case, refers to intentionally misleading content \cite{fallis2015disinformation}.   

A prominent case of a disinformation campaign are the efforts of the Russian government to control information during the Russia-Ukraine crisis \cite{pomerantsev2014}. One of the most important events during the crisis was the crash of Malaysian Airlines (MH17) flight on July 17, 2014. The plane crashed on its way from Amsterdam to Kuala Lumpur over Ukrainian territory, causing the death of 298 civilians. The event immediately led to the circulation of competing narratives about who was responsible for the crash (see Section \ref{sec:background}), with the two most prominent narratives being that the plane was either shot down by the Ukrainian military, or by Russian separatists in Ukraine supported by the Russian government \cite{oates2016}. The latter theory was confirmed by findings of an international investigation team. In this work, information that opposes these findings by promoting other theories about the crash is considered disinformation. When studying disinformation, however, it is important to acknowledge that our fact checkers (in this case the international investigation team) may be wrong, which is why we focus on both of the narratives in our study.

MH17 is a highly important case in the context of international relations, because the tragedy has not only increased Western, political pressure against Russia, but may also continue putting the government's global image at stake. In 2020, at least four individuals connected to the Russian separatist movement will face murder charges for their involvement in the MH17 crash \cite{harding_2019}, which is why one can expect the waves of disinformation about MH17 to continue spreading. The purpose of this work is to develop an approach that may help both practitioners and scholars of political science, international relations and political communication to detect and measure the scope of MH17-related disinformation.

Several studies analyse the framing of the crash and the spread of (dis)information about the event in terms of pro-Russian or pro-Ukrainian framing. These studies analyse information based on manually labeled content, such as television transcripts \cite{oates2016} or tweets \cite{IA, hjorth2019}. Restricting the analysis to manually labeled content ensures a high quality of annotations, but prohibits analysis from being extended to the full amount of available data. 
Another widely used method for classifying misleading content is to use distant annotations, for example to classify a tweet based on the domain of a URL that is shared by the tweet, or a hashtag that is contained in the tweet \cite{guess2019, gallacher2018, grinberg2019}. Often, this approach treats content from uncredible sources as misleading (e.g. misinformation, disinformation or fake news). This methods enables researchers to scale up the number of observations without having to evaluate the fact value of each piece of content from low-quality sources. However, the approach fails to address an important issue: Not all content from uncredible sources is necessarily misleading or false and not all content from credible sources is true. As often emphasized in the propaganda literature, established media outlets too are vulnerable to state-driven disinformation campaigns, even if they are regarded as credible sources \cite{jowett_propaganda_2014,taylor_munitions_2003, chomsky_manufacturing_1988}\footnote{The U.S. media coverage of weapons of mass destruction in Iraq stands as one of the most prominent examples of how generally credible sources can be exploited by state authorities.}.  

In order to scale annotations that go beyond metadata to larger datasets, Natural Language Processing (NLP) models can be used to automatically label text content. For example, several works developed classifiers for annotating text content with frame labels that can subsequently be used for large-scale content analysis \cite{Boydstun2014, Tsur2015, Card2015, johnson2017, Ji2017, Naderi17a, Field2018,  hartmann2019}. Similarly, automatically labeling attitudes expressed in text \cite{walker-etal-2012-stance,hasan-ng-2013-stance,Augenstein2016stance,journals/ipm/ZubiagaKLPLBCA18} can aid the analysis of disinformation and misinformation spread \cite{zubiaga2016analysing}. In this work, we examine to which extent such classifiers can be used to detect pro-Russian framing related to the MH17 crash, and to which extent classifier predictions can be relied on for analysing information flow on Twitter.

\paragraph{MH17 Related (Dis-)Information Flow on Twitter}
We focus our classification efforts on a Twitter dataset introduced in \citet{IA}, that was collected to investigate the flow of MH17-related information on Twitter, focusing on the question who is distributing (dis-)information. In their analysis, the authors found that citizens are active distributors, which contradicts the widely adopted view that the information campaign is only driven by the state and that citizens do not have an active role. \\
To arrive at this conclusion, the authors manually labeled a subset of the tweets in the dataset with pro-Russian/pro-Ukrainian frames and build a retweet network, which has Twitter users as nodes and edges between two nodes if a retweet occurred between the two associated users. An edge was considered as \textit{polarized} (either pro-Russian or pro-Ukrainian), if at least one retweet between the two users connected by the edge was pro-Russian/pro-Ukrainian. Then, the amount of polarized edges between users with different profiles (e.g. citizen, journalist, state organ) was computed.

Labeling more data via automatic classification (or computer-assisted annotation) of tweets could serve an analysis as the one presented in \citet{IA} in two ways. First, more edges could be labeled.\footnote{Only 26\% of the available tweets in \citet{IA}'s dataset are manually labeled.} Second, edges could be labeled with higher precision, i.e. by taking more tweets comprised by the edge into account. For example, one could decide to only label an edge as polarized if at least half of the retweets between the users were pro-Ukrainian/pro-Russian.

\paragraph{Contributions}
We evaluate different classifiers that predict frames for unlabeled tweets in \citet{IA}'s dataset, in order to increase the number of polarized edges in the retweet network derived from the data. This is challenging due to a skewed data distribution and the small amount of training data for the pro-Russian class. We try to combat the data sparsity using a data augmentation approach, 
but have to report a negative result as we find that data augmentation in this particular case does not improve classification results. While our best neural classifier clearly outperforms a hashtag-based baseline, generating high quality predictions for the pro-Russian class is difficult: In order to make predictions at a precision level of 80\%, recall has to be decreased to 23\%. Finally, we examine the applicability of the classifier for finding new polarized edges in a retweet network and show how, with manual filtering, the number of pro-Russian edges can be increased by 29\%. We make our code, trained models and predictions publicly available\footnote{\url{https://github.com/coastalcph/mh17}}.

\section{Competing Narratives about the MH17 Crash}\label{sec:background}
We briefly summarize the timeline around the crash of MH17 and some of the dominant narratives present in the dataset. On July 17, 2014, the MH17 flight crashed over Donetsk Oblast in Ukraine. The region was at that time part of an armed conflict between pro-Russian separatists and the Ukrainian military, one of the unrests following the Ukrainian revolution and the annexation of Crimea by the Russian government. The territory in which the plane fell down was controlled by pro-Russian separatists. 

Right after the crash, two main narratives were propagated: Western media claimed that the plane was shot down by pro-Russian separatists, whereas the Russian government claimed that the Ukrainian military was responsible. Two organisations were tasked with investigating the causes of the crash, the Dutch Safety Board (DSB) and the Dutch-led joint investigation team (JIT). Their final reports were released in October 2015 and September 2016, respectively, and conclude that the plane had been shot down by a missile launched by a BUK surface-to-air system. The BUK was stationed in an area controlled by pro-Russian separatists when the missile was launched, and had been transported there from Russia and returned to Russia after the incident. These findings are denied by the Russian government until now.
There are several other crash-related reports that are frequently mentioned throughout the dataset. One is a report by Almaz-Antey, the Russian company that manufactured the BUK, which rejects the DSB findings based on mismatch of technical evidence. Several reports backing up the Dutch findings were released by the investigative journalism website Bellingcat.\footnote{\url{https://www.bellingcat.com/}}

The crash also sparked the circulation of several alternative theories, many of them promoted in Russian media \cite{oates2016}, e.g. that the plane was downed by Ukrainian SU25 military jets, that the plane attack was meant to hit Putin’s plane that was allegedly traveling the same route earlier that day, and that the bodies found in the plane had already been dead before the crash.  

\section{Dataset}\label{sec:anno_guidelines}
For our classification experiments, we use the  MH17 Twitter dataset introduced by \citet{IA}, a dataset collected in order to study the flow of (dis)information about the MH17 plane crash on Twitter.
It contains tweets collected based on keyword search\footnote{These keywords were: MH17, Malazijskij [and] Boeing (in Russian), \#MH17, \#Pray4MH17, \#PrayforMH17. The dataset was collected using the Twitter \textit{Garden hose}, which means that it contains a 10\% of all tweets within the specified period that matched the search criterion.} that were posted between July 17, 2014 (the day of the plane crash) and December 9, 2016. 

\citet{IA} provide annotations for a subset of the English tweets contained in the dataset. A tweet is annotated with one of three classes that indicate the framing of the tweet with respect to responsibility for the plane crash. A tweet can either be \textit{pro-Russian} (Ukrainian authorities, NATO or EU countries are explicitly or implicitly held responsible, or the tweet states that Russia is not responsible), \textit{pro-Ukrainian} (the Russian Federation or Russian separatists in Ukraine are explicitly or implicitly held responsible, or the tweet states that Ukraine is not responsible) or \textit{neutral} (neither Ukraine nor Russia or any others are blamed). Example tweets for each category can be found in Table \ref{t:examples}. These examples illustrate that the framing annotations do not reflect general polarity, but polarity with respect to responsibility to the crash. For example, even though the last example in the table is in general pro-Ukrainian, as it displays the separatists in a bad light, the tweet does not focus on responsibility for the crash. Hence the it is labeled as neutral. 

Table \ref{t:anno_stats} shows the label distribution of the annotated portion of the data as well as the total amount of original tweets, and original tweets plus their retweets/duplicates in the network. A \textit{retweet} is a repost of another user's original tweet, indicated by a specific syntax (RT @username: ). We consider as \textit{duplicate} a tweet with text that is identical to an original tweet after preprocessing (see Section \ref{sec:preproc}). 
For our classification experiments, we exclusively consider original tweets, but model predictions can then be propagated to retweets and duplicates.

\begin{table}[t]
\centering
\resizebox{\linewidth}{!}{
\begin{tabular}{llrr}
 \toprule
 & Label & Original & All\\
 \toprule
\multirow{3}{*}{Labeled} & Pro-Russian & 512& 4,829 \\
& Pro-Ukrainian & 910& 12,343 \\
& Neutral& 6,923& 118,196 \\
\midrule
Unlabeled  & - & 192,003&377,679\\
\bottomrule
Total & - &200,348 & 513,047\\

\end{tabular}}
\caption{Label distribution and dataset sizes. Tweets are considered \textit{original} if their preprocessed text is unique. \textit{All} tweets comprise original tweets, retweets and duplicates.}\label{t:anno_stats}
\end{table}




\begin{table*}[t]
\centering
\resizebox{\linewidth}{!}{
\begin{tabular}{l l}
\toprule
Label & Example tweet \\
\midrule
\multirow{4}{*}{Pro-Ukrainian} & Video - Missile that downed MH17 ’was brought in from Russia’ @peterlane5news\\
 & RT @mashable: Ukraine: Audio recordings show pro-Russian rebels tried to hide \#MH17 black boxes. \\
 & Russia Calls For New Probe Into MH17 Crash. Russia needs to say, ok we fucked up.. Rather than play games\\
 & @IamMH17 STOP LYING! You have ZERO PROOF to falsely blame UKR for \#MH17 atrocity. You will need to apologize.\\
\midrule
\multirow{2}{*}{Pro-Russian} &Why the USA and Ukraine, NOT Russia, were probably behind the shooting down of flight \#MH17 \\
 & RT @Bayard\_1967: UKRAINE Eyewitness Confirm Military Jet Flew Besides MH17 Airliner: BBC ...\\ 
 & RT @GrahamWP\_UK: Just read through \#MH17 @bellingcat report, what to say - written by frauds, believed by the gullible. Just that.\\
 \midrule
 \multirow{2}{*}{Neutral} & \#PrayForMH17 :( \\
 & RT @deserto\_fox: Russian terrorist stole wedding ring from dead passenger \#MH17 \\
 \bottomrule
\end{tabular}}
\caption{Example tweets for each of the three classes.}\label{t:examples}
\end{table*}

\section{Classification Models}\label{sec:models}
For our classification experiments, we compare three classifiers, a hashtag-based baseline, a logistic regression classifier and a convolutional neural network (CNN).

\paragraph{Hashtag-Based Baseline}
Hashtags are often used as a means to assess the content of a tweet \cite{efron2010hashtag, godin2013using, dhingra-etal-2016-tweet2vec}. We identify hashtags indicative of a class in the annotated dataset using the pointwise mutual information (pmi) between a hashtag $hs$ and a class $c$, which is defined as 
\begin{equation}
    \text{pmi}(hs, c) = \text{log}\dfrac{\text{p}(hs, c)}{\text{p}(hs)\text{\ p}(c)}
\end{equation}
We then predict the class for unseen tweets as the class that has the highest pmi score for the hashtags contained in the tweet.
Tweets without hashtag (5\% of the tweets in the development set) or with multiple hashtags leading to conflicting predictions (5\% of the tweets in the development set) are labeled randomly. We refer to to this baseline as \textsc{hs\_pmi}.

\paragraph{Logistic Regression Classifier}
As non-neural baseline we use a logistic regression model.\footnote{As non-neural alternative, we also experimented with SVMs. These showed inferior performance to the regression model.} We compute input representations for tweets as the average over pre-trained word embedding vectors for all words in the tweet. We use fasttext embeddings \cite{fasttext} that were pre-trained on Wikipedia.\footnote{In particular, with cross-lingual experiments in mind (see Section \ref{sec:crosslingual}), we used embeddings that are pre-aligned between languages available here \url{https://fasttext.cc/docs/en/aligned-vectors.html}}

\paragraph{Convolutional Neural Network Classifier}
As neural classification model, we use a convolutional neural network (CNN) \cite{kim2014convolutional}, which has previously shown good results for tweet classification \cite{dos-santos-gatti-2014-deep, dhingra-etal-2016-tweet2vec}.\footnote{We also ran intitial experiments with recurrent neural networks (RNNs), but found that results were comparable with those achieved by the CNN architecture, which runs considerably faster.} The model performs 1d convolutions over a sequence of word embeddings. We use the same pre-trained fasttext embeddings as for the logistic regression model. We use a model with one convolutional layer and a relu activation function, and one max pooling layer. The number of filters is 100 and the filter size is set to 4.

\section{Experimental Setup}
We evaluate the classification models using 10-fold cross validation, i.e. we produce 10 different datasplits by randomly sampling 60\% of the data for training, 20\% for development and 20\% for testing. For each fold, we train each of the models described in Section \ref{sec:models} on the training set and measure performance on the test set. For the \textsc{CNN} and \textsc{LogReg} models, we upsample the training examples such that each class has as many instances as the largest class (Neutral). The final reported scores are averages over the 10 splits.\footnote{We train with the same hyperparameters on all splits, these hyperparameters were chosen according to the best macro f score averaged over 3 runs with different random seeds on \textit{one} of the splits.}

\subsection{Tweet Preprocessing}\label{sec:preproc}
Before embedding the tweets, we replace urls, retweet syntax (RT @user\_name: ) and @mentions (@user\_name) by placeholders. We lowercase all text and tokenize sentences using the StandfordNLP pipeline \cite{stanfordnlp}. If a tweet contains multiple sentences, these are concatenated. Finally, we remove all tokens that contain non-alphanumeric symbols (except for dashes and hashtags) and strip the hashtags from each token, in order to increase the number of words that are represented by a pre-trained word embedding.

\subsection{Evaluation Metrics}
We report performance as F1-scores, which is the harmonic mean between precision and recall. As the class distribution is highly skewed and we are mainly interested in accurately classifying the classes with low support (pro-Russian and pro-Ukrainian),  we report macro-averages over the classes. In addition to F1-scores, we report the area under the precision-recall curve (AUC).\footnote{The AUC is computed according to the trapezoidal rule, as implemented in the sklearn package \cite{scikit-learn}} We compute an AUC score for each class by converting the classification task into a one-vs-all classification task.

\begin{table*}[t]
\centering
\begin{tabular}{l|rr|rr|rr| rr}
\toprule
&\multicolumn{2}{c}{Macro-avg} & \multicolumn{2}{c}{Pro-Russian} & \multicolumn{2}{c}{Pro-Ukrainian}&
\multicolumn{2}{c}{Neutral}\\
Model & F1   & AUC & F1  & AUC & F1  & AUC & F1  & AUC \\
\midrule
\textsc{Random} & 0.25& - &0.10 &-&0.16 & - & 0.47& - \\
\textsc{hs\_pmi} & 0.25&- &0.10&- &0.16& - & 0.48&   -\\
\textsc{LogReg} & 0.59 & 0.53 & 0.38 &0.34& 0.51 & 0.41& 0.88 & 0.86\\
\textsc{CNN} & \textbf{0.69}& \textbf{0.71}& \textbf{0.55}& \textbf{0.57}& \textbf{0.59}& \textbf{0.60}& \textbf{0.93} & \textbf{0.94}\\
\bottomrule
\end{tabular}
\caption{Classification results on the English MH17 dataset measured as F1 and area under the precision-recall curve (AUC).}\label{t:results}

\end{table*}

\begin{figure*}[t]

\includegraphics[width=.25\linewidth]{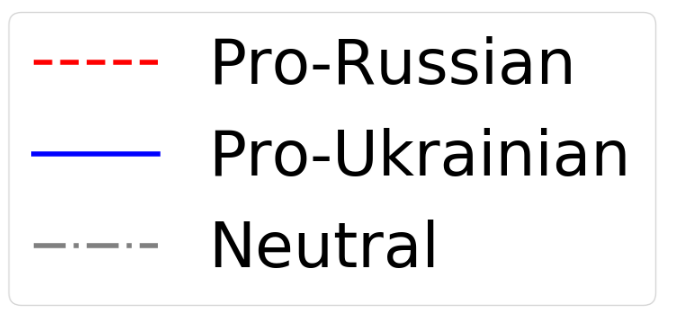}
\centering
\medskip
\includegraphics[width=.5\linewidth]{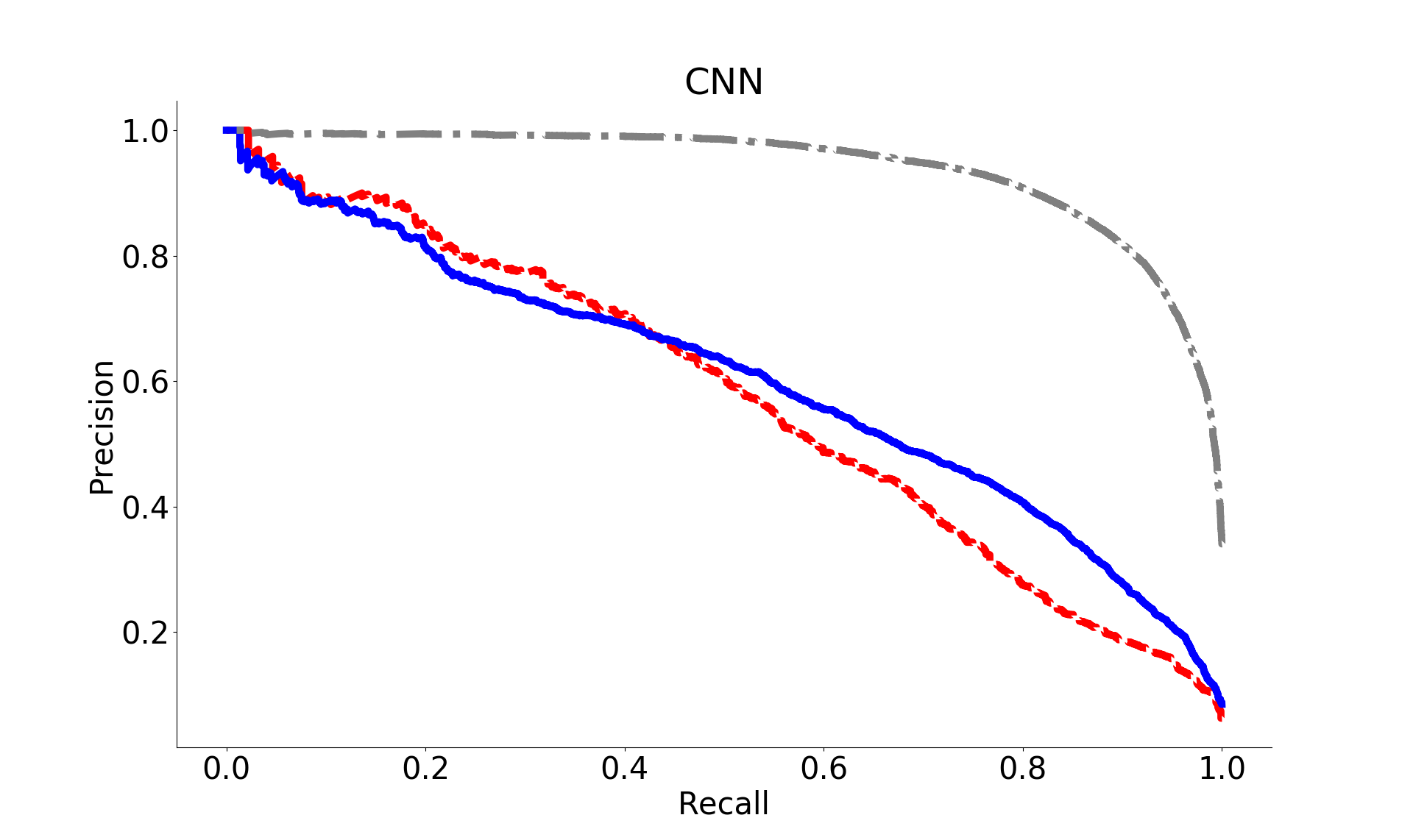}\includegraphics[width=.5\linewidth]{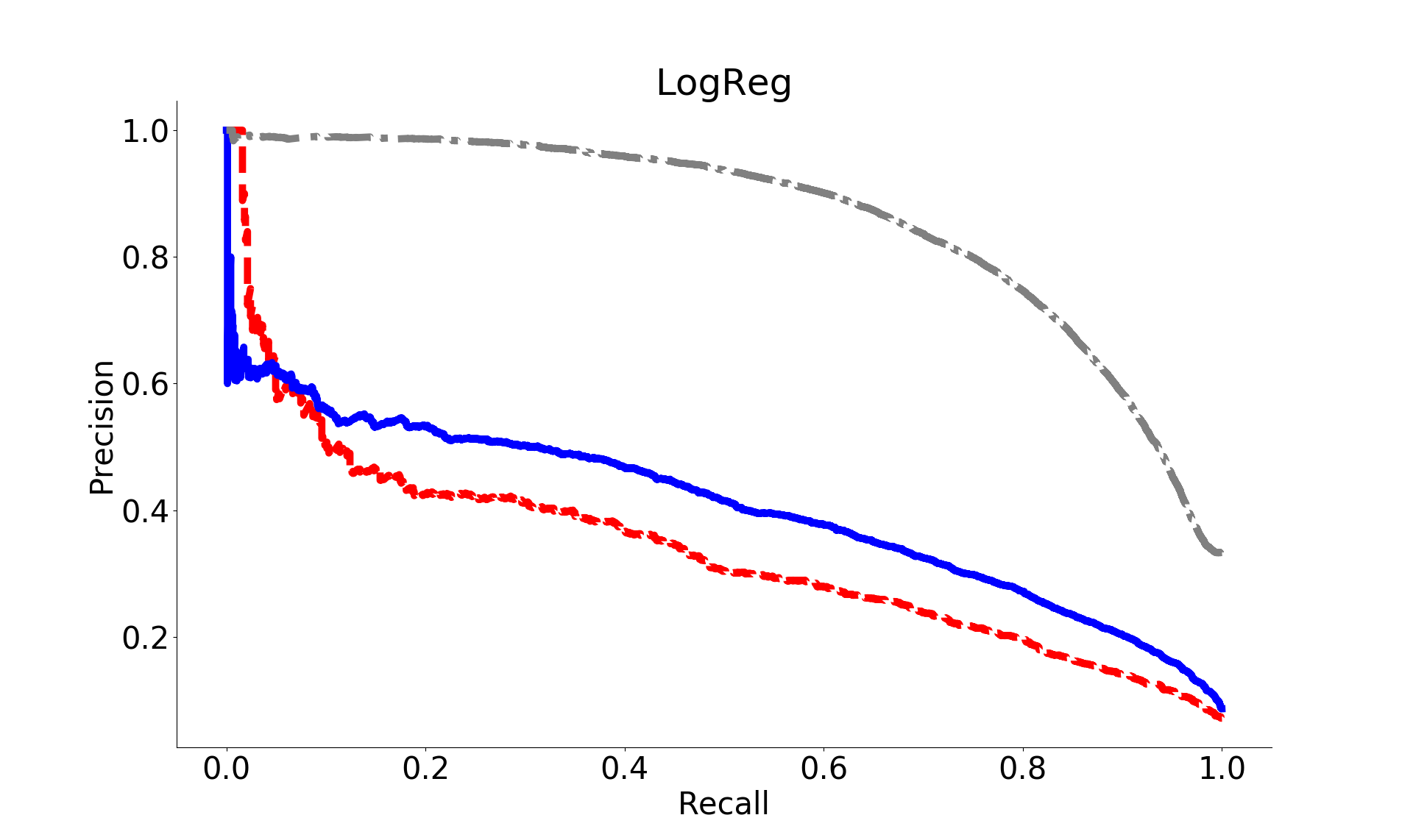}
\includegraphics[width=\linewidth]{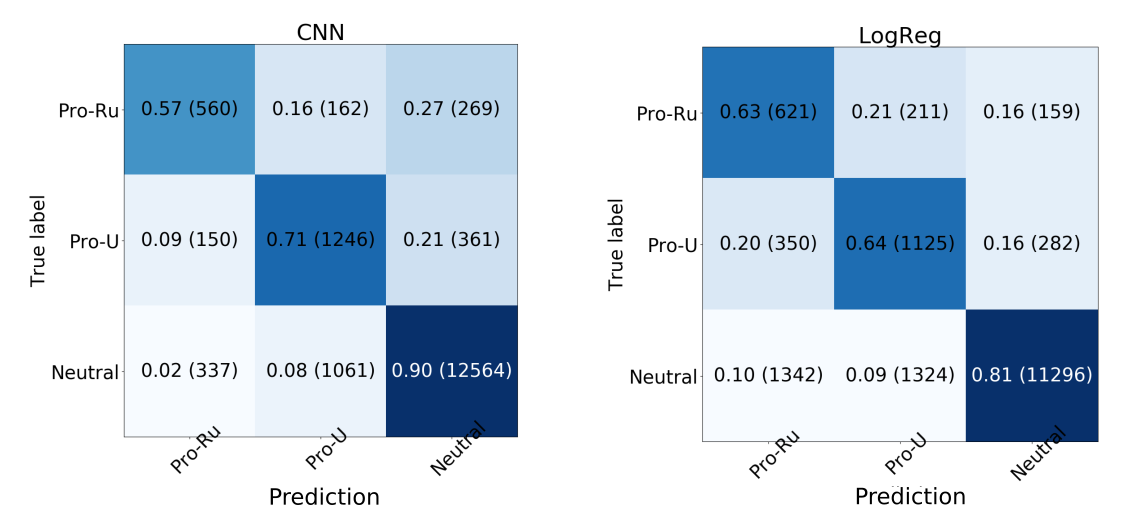}
\caption{Confusion matrices for the CNN (left) and the logistic regression model (right). The y-axis shows the true label while the x-axis shows the model prediction.}\label{f:cm}
\end{figure*}

\section{Results}
The results of our classification experiments are presented in Table \ref{t:results}. Figure \ref{f:cm} shows the per-class precision-recall curves for the \textsc{LogReg} and \textsc{CNN} models as well as the confusion matrices between classes.\footnote{Both the precision-recall curves and the confusion matrices were computed by concatenating the test sets of all 10 datasplits} 

\paragraph{Comparison Between Models}
We observe that the hashtag baseline performs poorly and does not improve over the random baseline. The \textsc{CNN} classifier outperforms the baselines as well as the \textsc{LogReg} model. It shows the highest improvement over the \textsc{LogReg} for the pro-Russian class. Looking at the confusion matrices, we observe that for the \textsc{LogReg} model, the fraction of True Positives is equal between the pro-Russian and the pro-Ukrainian class. The \textsc{CNN} model produces a higher amount of correct predictions for the pro-Ukrainian than for the pro-Russian class. The absolute number of pro-Russian True Positives is lower for the \textsc{CNN}, but so is in return the amount of misclassifications between the pro-Russian and pro-Ukrainian class. 

\paragraph{Per-Class Performance}
With respect to the per class performance, we observe a similar trend across models, which is that the models perform best for the neutral class, whereas performance is lower for the pro-Ukrainian and pro-Russian classes. All models perform worst on the pro-Russian class, which might be due to the fact that it is the class with the fewest instances in the dataset.

Considering these results, we conclude that the \textsc{CNN} is the best performing model and also the classifier that best serves our goals, as we want to produce accurate predictions for the pro-Russian and pro-Ukrainian class without confusing between them. Even though the \textsc{CNN} can improve over the other models, the classification performance for the pro-Russian and pro-Ukrainian class is rather low. One obvious reason for this might be the small amount of training data, in particular for the pro-Russian class.  

In the following, we briefly report a negative result on an attempt to combat the data sparseness with cross-lingual transfer. We then perform an error analysis on the \textsc{CNN} classifications to shed light on the difficulties of the task.

\section{Data Augmentation Experiments using Cross-Lingual Transfer}\label{sec:crosslingual}
The annotations in the MH17 dataset are highly imbalanced, with as few as 512 annotated examples for the pro-Russian class. As the annotated examples were sampled from the dataset at random, we assume that there are only few tweets with pro-Russian stance in the dataset. This observation is in line with studies that showed that the amount of disinformation on Twitter is in fact small \cite{guess2019, grinberg2019}. In order to find more pro-Russian training examples, we turn to a resource that we expect to contain large amounts of pro-Russian (dis)information. The \textit{Elections integrity dataset}\footnote{\url{https://about.twitter.com/en_us/values/elections-integrity.html\#data}} was released by Twitter in 2018 and contains the tweets and account information for 3,841 accounts that are believed to be Russian trolls financed by the Russian government. While most tweets posted after late 2014 are in English language and focus on topics around the US elections, the earlier tweets in the dataset are primarily in Russian language and focus on the Ukraine crisis \cite{howard2018ira}. One feature of the dataset observed by \citet{howard2018ira} is that several hashtags show high peakedness \cite{etling}, i.e. they are posted with high frequency but only during short intervals, while others are persistent during time.

We find two hashtags in the Elections integrity dataset with high peakedness that were exclusively posted within 2 days after the MH17 crash and that seem to be pro-Russian in the context of responsibility for the MH17 crash: \begin{otherlanguage*}{russian}
\#КиевСкажиПравду
\end{otherlanguage*} (\textit{Kiew tell the truth}) and \begin{otherlanguage*}{russian}
\#Киевсбилбоинг
\end{otherlanguage*} (\textit{Kiew made the plane go down}). We collect all tweets with these two hashtags, resulting in 9,809 Russian tweets that we try to use as additional training data for the pro-Russian class in the MH17 dataset. We experiment with cross-lingual transfer by embedding tweets via aligned English and Russian word embeddings.\footnote{We use two sets of monolingual fasttext embeddings trained on Wikipedia \cite{fasttext} that were aligned relying on a seed lexicon of 5000 words via the RCSLS method \cite{rcsls}} However, so far results for the cross-lingual models do not improve over the \textsc{CNN} model trained on only English data. This might be due to the fact that the additional Russian tweets rather contain a general pro-Russian frame than specifically talking about the crash, but needs further investigation.


\begin{table*}[t]
\centering
\resizebox{\linewidth}{!}{
\begin{tabular}{p{1cm}p{1cm}p{2cm}l p{15cm}}
\toprule
Error cat. & True class & Model prediction & \parbox{0.05\linewidth}{\vspace*{6pt} id} & \parbox{0.05\linewidth}{\vspace*{6pt}Tweet}\\
\midrule
 \multirow{6}{*}{\parbox{1\linewidth}{\vspace*{79pt} I}}   &\multirow{3}{*}{\parbox{1\linewidth}{\vspace*{1cm}Pro-U}}  & \multirow{3}{*}{\parbox{1\linewidth}{\vspace*{1cm}Pro-R}} & a)& RT @ChadPergram: Hill intel sources say Russia has the capability to potentially shoot down a \#MH17 but not Ukraine.\\
  & & & b) & RT @C4ADS: .@bellingcat's new report says \#Russia used fake evidence for \#MH17 case to blame \#Ukraine URL\\
    & & & c) & The international investigation blames Russia for MH17 crash URL \#KievReporter \#MH17 \#Russia \#terror \#Ukraine \#news \#war\\
 \cmidrule{2-5}
 &\multirow{3}{*}{\parbox{1\linewidth}{\vspace*{14pt}Pro-R}}  & \multirow{3}{*}{\parbox{1\linewidth}{\vspace*{14pt}Pro-U}} & d) & RT @RT\_com: BREAKING: No evidence of direct Russian link to \#MH17 - US URL URL\\
 & &  & e) & RT @truthhonour: Yes Washington was behind Eukraine jets that shot down MH17 as pretext to conflict with Russia. No secrets there\\
 & &  & f) & Ukraine Media Falsely Claim Dutch Prosecutors Accuse Russia of Downing MH17: Dutch prosecutors de URL \#MH17 \#alert\\

   \midrule
   \multirow{7}{*}{\parbox{1\linewidth}{\vspace*{71pt} II}}  &\multirow{3}{*}{\parbox{1\linewidth}{\vspace*{1cm}Pro-U}}  & \multirow{3}{*}{\parbox{1\linewidth}{\vspace*{1cm}Pro-R}} & g)& @Werteverwalter @Ian56789 @ClarkeMicah no SU-25 re \#MH17 believer has ever been able to explain it,facts always get in their way\\
  & & & h) & Rebel theories on \#MH17 "total nonsense", Ukrainian Amb to U.S. Olexander Motsyk interviewed by @jaketapper via @cnn\\
    & & & i) & Ukrainian Pres. says it's false "@cnnbrk: Russia says records indicate Ukrainian warplane was flying within 5 km of \#MH17 on day of crash.
\\
 \cmidrule{2-5}
 &\multirow{4}{*}{\parbox{1\linewidth}{\vspace*{43pt}Pro-R}}  & \multirow{4}{*}{\parbox{1\linewidth}{\vspace*{43pt}Pro-U}} & j)&  Russia has released some solid evidence to contradict @EliotHiggins + @bellingcat's \#MH17 report. http://t.co/3leYfSoLJ3\\
   & & & k) & RT @masamikuramoto: @MJoyce2244 The jets were seen by Russian military radar and Ukrainian eyewitnesses. \#MH17 @Fossibilities @irina\\
& &  & l)& RT @katehodal: Pro-Russia separatist says \#MH17 bodies "weren't fresh" when found in Ukraine field,suggesting already dead b4takeoff\\
    & & & m) & RT @NinaByzantina: \#MH17 redux: 1) \#Kolomoisky admits involvement URL 2) gets \$1.8B of \#Ukraine's bailout funds\\

  \midrule
   \multirow{6}{*}{\parbox{1\linewidth}{\vspace*{34pt} III}}  &\multirow{3}{*}{\parbox{1\linewidth}{\vspace*{14pt}Pro-U}}  & \multirow{3}{*}{\parbox{1\linewidth}{\vspace*{14pt}Pro-R}} & n)& \#Russia again claiming that \#MH17 was shot down by air-to-air missile, which of course wasn't russian-made. \#LOL URL\\
      &  &  & o)& RT @20committee: New Moscow line is \#MH17 was shot down by a Ukrainian fighter. With an LGBT pilot, no doubt.\\
 \cmidrule{2-5}
  &\multirow{3}{*}{\parbox{1\linewidth}{\vspace*{34pt}Pro-R}}  & \multirow{3}{*}{\parbox{1\linewidth}{\vspace*{34pt}Pro-U}} & q)& RT @merahza: If you believe the pro Russia rebels shot \#MH17 then you'll believe Justine Bieber is the next US President and that Coke is a \\
   & & & q) & So what @AC360 is implying is that \#US imposed sanctions on \#Russia, so in turn they shot down a \#Malaysia jet carrying \#Dutch people? \#MH17\\
   & & & r) &  RT @GrahamWP\_UK: \#MH17  1. A man on sofa watching YouTube thinks it was a 'separatist BUK'.  2. Man on site for over 25 hours doesn't.\\

  \bottomrule

\end{tabular}}
\caption{Examples for the different error categories. Error category I are cases where the correct class can easily be inferred from the text. For error category II, the correct class can be inferred from the text with event-specific knowledge. For error category III, it is necessary to resolve humour/satire in order to infer the intended meaning that the speaker wants to communicate. }\label{t:error_analysis}
\end{table*}

\section{Error Analysis}
In order to integrate automatically labeled examples into a network analysis that studies the flow of polarized information in the network, we need to produce high precision predictions for the pro-Russian and the pro-Ukrainian class. Polarized tweets that are incorrectly classified as neutral will hurt an analysis much less than neutral tweets that are erroneously classified as pro-Russian or pro-Ukrainian. However, the worst type of confusion is between the pro-Russian and pro-Ukrainian class. In order to gain insights into why these confusions happen, we manually inspect incorrectly predicted examples that are confused between the pro-Russian and pro-Ukrainian class. We analyse the misclassifications in the development set of all 10 runs, which results in 73 False Positives of pro-Ukrainian tweets being classified as pro-Russian (referred to as \textit{pro-Russian False Positives}), and 88 False Positives of pro-Russian tweets being classified as pro-Ukrainian (referred to as \textit{pro-Ukrainian False Positives}). We can identify three main cases for which the model produces an error: 
\begin{enumerate}
    \item the correct class can be directly inferred from the text content easily, even without background knowledge
    \item the correct class can be inferred from the text content, given that event-specific knowledge is provided
    \item the correct class can be inferred from the text content if the text is interpreted correctly
\end{enumerate}
For the pro-Russian False Positives, we find that 42\% of the errors are category I and II errors, respectively, and 15\% of category III. For the pro-Ukrainian False Positives, we find 48\% category I errors, 33\% category II errors and and 13\% category III errors. Table \ref{t:error_analysis} presents examples for each of the error categories in both sets which we will discuss in the following. 

\paragraph{Category I Errors}
Category I errors could easily be classified by humans following the annotation guidelines (see Section \ref{sec:anno_guidelines}). One difficulty can be seen in example f). Even though no background knowledge is needed to interpret the content, interpretation is difficult because of the convoluted syntax of the tweet. For the other examples it is unclear why the model would have difficulties with classifying them.

\paragraph{Category II Errors}
Category II errors can only be classified with event-specific background knowledge. Examples g), i) and k) relate to the theory that a Ukrainian SU25 fighter jet shot down the plane in air. Correct interpretation of these tweets depends on knowledge about the SU25 fighter jet. In order to correctly interpret example j) as pro-Russian, it has to be known that the bellingcat report is pro-Ukrainian. Example l) relates to the theory that the shoot down was a false flag operation run by Western countries and the bodies in the plane were already dead before the crash. In order to correctly interpret example m), the identity of \textit{Kolomoisky} has to be known. He is an anti-separatist Ukrainian billionaire, hence his involvement points to the Ukrainian government being responsible for the crash.

\paragraph{Category III Errors}
Category III errors occur for examples that can only be classified by correctly interpreting the tweet authors' intention. Interpretation is difficult due to phenomena such as irony as in examples n) and o). While the irony is indicated in example n) through the use of the hashtag \textit{\#LOL}, there is no explicit indication in example o).\\
Interpretation of example q) is conditioned on world knowledge as well as the understanding of the speakers beliefs. Example r) is pro-Russian as it questions the validity of the assumption AC360 is making, but we only know that because we know that the assumption is absurd. Example s) requires to evaluate that the speaker thinks people on site are trusted more than people at home.

From the error analysis, we conclude that category I errors need further investigation, as here the model makes mistakes on seemingly easy instances. This might be due to the model not being able to correctly represent Twitter specific language or unknown words, such as \textit{Eukraine} in example e). Category II and III errors are harder to avoid and could be improved by applying reasoning \cite{reasoning} or irony detection methods \cite{irony}.

\begin{figure*}[tb]
\resizebox{\linewidth}{!}{
\centering
\includegraphics[]{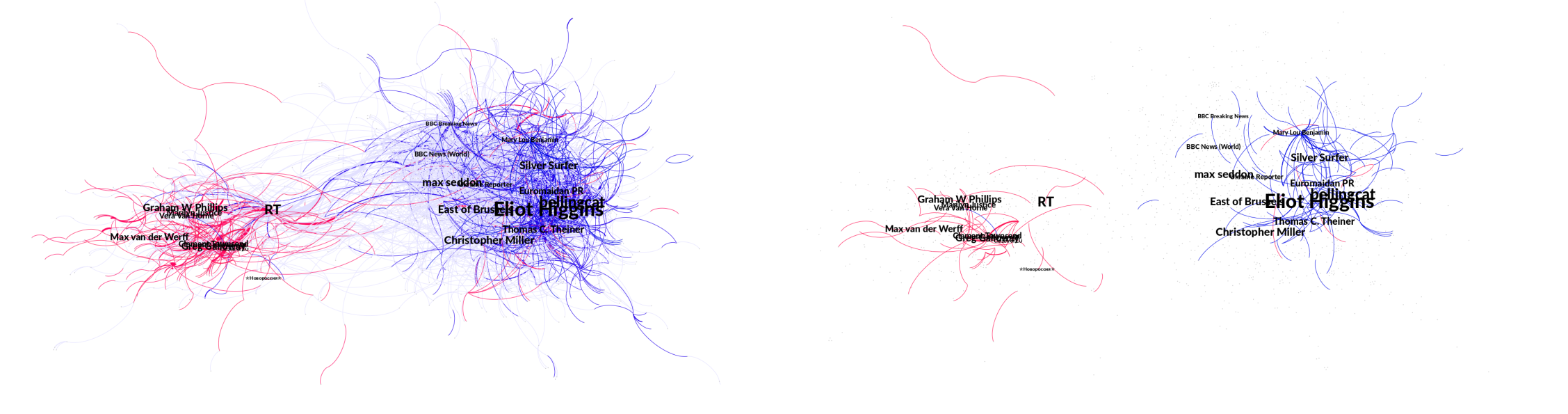} 
}
\caption{The left plot shows the original k10 retweet network as computed by \citet{IA} together with the new edges that were added after manually re-annotating the classifier predictions. The right plot only visualizes the new edges that we could add by filtering the classifier predictions. Pro-Russian edges are colored in red, pro-Ukrainian edges are colored in dark blue and neutral edges are colored in grey. Both plots were made using The Force Atlas 2 layout in gephi \cite{gephi}.}\label{f:orig_k10}
\end{figure*}

\section{Integrating Automatic Predictions into the Retweet Network}
Finally, we apply the \textsc{CNN} classifier to label new edges in \citet{IA}'s retweet network, which is shown in Figure \ref{f:orig_k10}. The retweet network is a graph that contains users as nodes and an edge between two users if the users are retweeting each other.\footnote{\citet{IA} use the k10 core of the network, which is the maximal subset of nodes and edges, such that all included nodes are connected to at least k other nodes \cite{seidman1983}, i.e. all users in the network have interacted with at least 10 other users.} In order to track the flow of polarized information, \citet{IA} label an edge as polarized if at least one tweet contained in the edge was manually annotated as pro-Russian or pro-Ukrainian. While the network shows a clear polarization, only a small subset of the edges present in the network are labeled (see Table \ref{t:k10_stats}). 

Automatic polarity prediction of tweets can help the analysis in two ways. Either, we can label a previously unlabeled edge, or we can verify/confirm the manual labeling of an edge, by labeling additional tweets that are comprised in the edge.


\subsection{Predicting Polarized Edges}
In order to get high precision predictions for unlabeled tweets, we choose the probability thresholds for predicting a pro-Russian or pro-Ukrainian tweet such that the classifier would achieve 80\% precision on the test splits (recall at this precision level is 23\%). Table \ref{t:k10_stats} shows the amount of polarized edges we can predict at this precision level. Upon manual inspection, we however find that the quality of predictions is lower than estimated. Hence, we manually re-annotate the pro-Russian and pro-Ukrainian predictions according to the official annotation guidelines used by \cite{IA}. This way, we can label 77 new pro-Russian edges by looking at 415 tweets, which means that 19\% of the candidates are hits.  For the pro-Ukrainian class, we can label 110 new edges by looking at 611 tweets (18\% hits). Hence even though the quality of the classifier predictions is too low to be integrated into the network analysis right away, the classifier drastically facilitates the annotation process for human annotators compared to annotating unfiltered tweets (from the original labels we infer that for unfiltered tweets, only 6\% are hits for the pro-Russian class, and 11\% for the pro-Ukrainian class).

\begin{table}[t]
\centering
\resizebox{\linewidth}{!}{
\begin{tabular}{p{3.5cm}rrrrr}
\toprule
& Pro-R & Pro-U & Neutral & Total  \\
\midrule
\# labeled edges in k10 & 270& 678& 2193& 3141 \\
\# candidate edges & 349& 488 & - & 873\\
\# added after filtering predictions & \textbf{77} & \textbf{110}& -& 187\\
\end{tabular}}
\caption{Number of labeled edges in the k10 network before and after augmentation with predicted labels. Candidates are previously unlabeled edges for which the model makes a confident prediction. The total number of edges in the network is 24,602.}\label{t:k10_stats}
\end{table}

\section{Conclusion}
In this work, we investigated the usefulness of text classifiers to detect pro-Russian and pro-Ukrainian framing in tweets related to the MH17 crash, and to which extent classifier predictions can be relied on for producing high quality annotations. From our classification experiments, we conclude that the real-world applicability of text classifiers for labeling polarized tweets in a retweet network is restricted to pre-filtering tweets for manual annotation. However, if used as a filter, the classifier can significantly speed up the annotation process, making large-scale content analysis more feasible. 

\section*{Acknowledgements}
We thank the anonymous reviewers for their helpful comments. The research was carried out as part of the ‘Digital Disinformation’ project, which was directed by Rebecca Adler-Nissen and funded by the Carlsberg Foundation (project number CF16-0012).

\bibliography{emnlp-ijcnlp-2019}
\bibliographystyle{acl_natbib}

\appendix

\end{document}